%% file: erk.tex
\documentclass[a4paper]{article}
\usepackage{cite}

\usepackage[utf8]{inputenc}
\usepackage{erk}
\usepackage{times}
\usepackage{graphicx}
\usepackage[top=22.5mm, bottom=22.5mm, left=22.5mm, right=22.5mm]{geometry}

\usepackage{multirow}
\usepackage{booktabs} % For professional looking tables

\usepackage{amsmath}
\usepackage{amssymb}
\usepackage{amsthm}
\usepackage{amsfonts}  % For \tensor

\newcommand{\R}{\mathbb{R}}

\newcommand{\N}{\mathbb{N}}
\newcommand{\D}{D}
\newcommand{\I}{\mathcal{I}}
\newcommand{\tensor}[1]{\underline{#1}}

\usepackage[slovene,english]{babel}

\def\footnotemark{}%  to avoid footnote on cover page

\begin{document}
%make title
%\title{Primerjalni pregled algoritma federativnega učenja s samorazvijajočim se mehkim sistemom}%
\title{Federativno učenje na podlagi samorazvijajočega se Gaussovega rojenja}%

\thanks{Prvi avtor se zahvaljuje Javni agenciji za znanstvenoraziskovalno in inovacijsko dejavnost Republike Slovenije (Slovenian Research and Innovation Agency -- ARIS) za finančno podporo, projekt P2-0219.}%

\author{Miha Ožbot$^{1}$, Igor Škrjanc$^{2}$} % use ^1, ^2 for author(s) from different institutions

\affiliation{$^{1,2}$Fakulteta za elektrotehniko, Univerza v Ljubljani, Slovenija}%%
\email{$^{1}$miha.ozbot@fe.uni-lj.si, $^{2}$igor.skrjanc@fe.uni-lj.si}%%
\maketitle

\begin{abstract}{Federated Learning based on Self-Evolving Gaussian Clustering}
%{Benchmark Examination Of A Federated Learning Algorithm With Self-Evolving Fuzzy System}
%
In this study, we present an Evolving Fuzzy System within the context of Federated Learning, which adapts dynamically with the addition of new clusters and therefore does not require the number of clusters to be selected apriori. Unlike traditional methods, Federated Learning allows models to be trained locally on clients' devices, sharing only the model parameters with a central server instead of the data. Our method, implemented using PyTorch, was tested on clustering and classification tasks. The results show that our approach outperforms established classification methods on several well-known UCI datasets. While computationally intensive due to overlap condition calculations, the proposed method demonstrates significant advantages in decentralized data processing.
\end{abstract}
\selectlanguage{slovene}
\section{Uvod}
\par Zaradi vse večje zaskrbljenosti glede zasebnosti in varnosti podatkov se je povečala potreba po metodologijah strojnega učenja, katerih cilj je decentralizacija postopka učenja modelov. Federativno učenje (\textit{Federated Learning} FL)~\cite{McMahan_Moore_Ramage_Hampson_Arcas_2023} predstavlja privlačno rešitev v scenarijih, kjer podatkov ni mogoče zbirati na centralnem strežniku zaradi skrbi za zasebnost, regulativnih omejitev ali velike količine generiranih podatkov. Namesto da bi se vsi podatki zbirali na osrednjem strežniku, se modeli učijo pri lastnikih podatkov~\cite{Konecnx_McMahan_2017}, v osrednji strežnik pa se za združevanje pošljejo le posodobitve lokalnih modelov, ne pa tudi izvorni podatki.
\par Primarni doprinos te raziskave je prilagoditev samorazvijajočih se mehkih sistemov (\textit{Evolving Fuzzy Systems} -- EFS)~\cite{Angelov_Xiaowei_Zhou_2008, Pratama_Anavatti_Joo_Lughofer_2015, Skrjanc_2020,Lughofer_2022} za uporabo v federativnem učenju. Ti ponujajo rešitve za več ključnih izzivov, s katerimi se srečamo pri FL. Glavni izziv pri nenadzorovanem rojenju je potreba po vnaprejšnji določitvi števila rojev, kar je še posebej težavno pri FL, kjer je treba to odločitev sprejeti za vsakega lastnika posebej~\cite{Stallmann_Wilbik_2022}. Nasprotno pa samorazvijajoče se mehko rojenje dinamično dodaja nove roje, s čimer se tej težavi popolnoma izogne~\cite{Skrjanc_2020}. Samorazvijajoči se sistemi nimajo težav s počasno konvergenco v primeru neodvisno in identično porazdeljenih (\textit{Non-Independent and Identically Distributed} --  Non-IID) podatkov~\cite{Li_Huang_Yang_Wang_Zhang_2020}, saj so roji lokalno veljavni – če se roji več lastnikov prekrivajo, se združijo, sicer pa ostanejo ločeni. Samorazvijajoči se sistemi dosegajo dobre rezultate pri enkratnem prehodu čez podatke, saj so zasnovani za spretno učenje v odprti zanki. Lokalne modele je mogoče združiti v globalni model z mehanizmi združevanja rojev, ki se uporabljajo v samorazvijajočih se mehkih sistemih. Poleg tega je temeljna lastnost mehkih sistemov njihova pregledna struktura, članstvo na podlagi rojev, pravila če-potem in lokalna linearnost pravil, kar omogoča interpretacijo rezultatov sklepanja modela.
\begin{figure}
    \centering
    \small
    \includegraphics[width=0.9\linewidth]{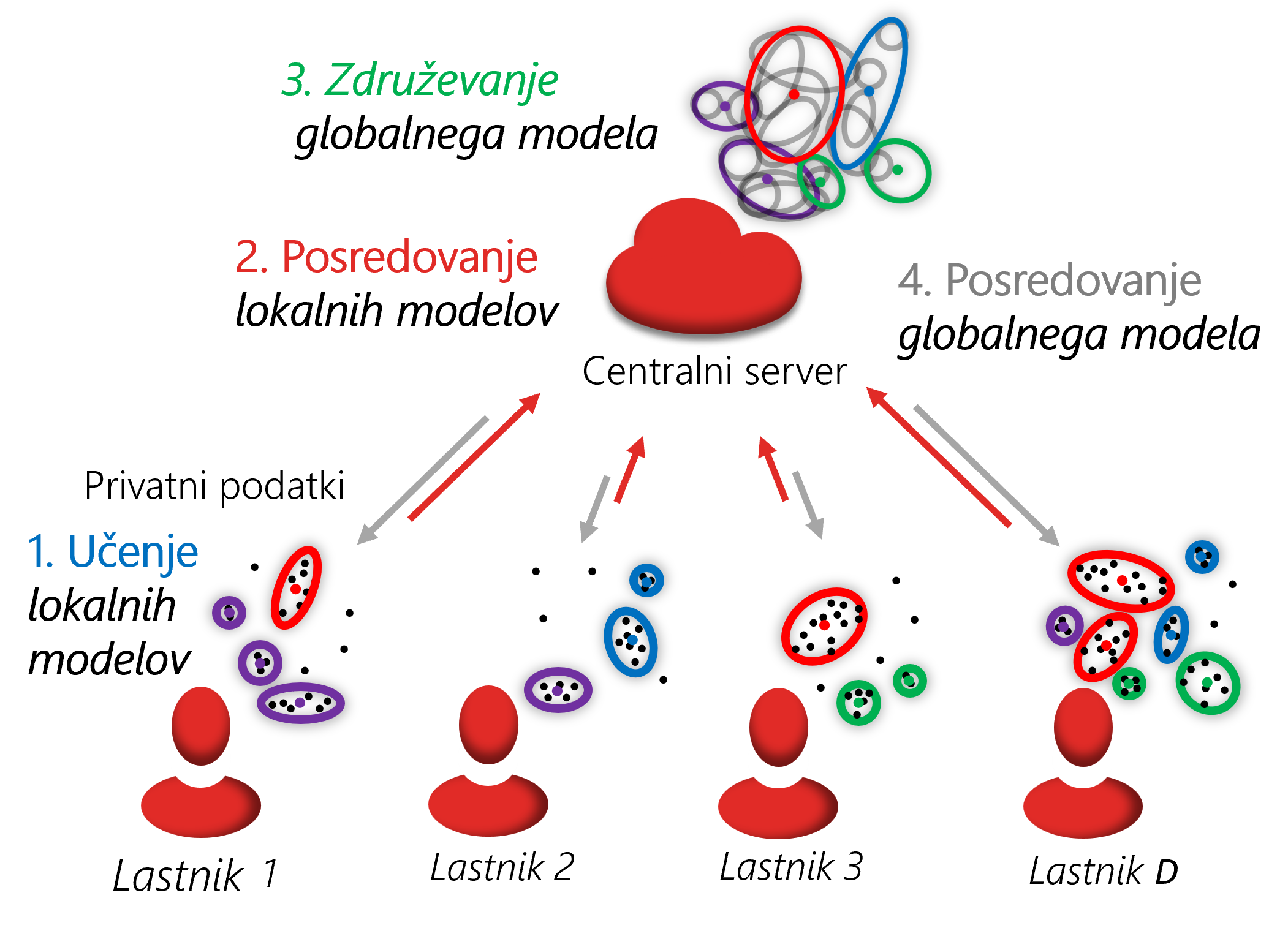}
    \caption{Predlagan algoritem federativnega učenja: (1) Vsak lastnik podatkov nauči svoj lokalni model na zasebnih podatkih. (2) Parametri teh lokalnih modelov se pošljejo na centralni strežnik. (3) Na strežniku se lokalni modeli združijo v en globalni model. (4) Ta združeni globalni model se nato prenese nazaj k vsakemu lokalnemu vozlišču.}
    \label{fig:Shema}
\end{figure}
\par Predlagan algoritem je predstavljen na sliki~\ref{fig:Shema}. Osredotočamo se na samorazvijajoče se mehke klasifikatorje (\textit{Evolving Fuzzy Classifier} -- EFC), ki so bili predlagani v~\cite{Angelov_Xiaowei_Zhou_2008} z družino klasifikatorjev eClass. Avtorji predlagajo dve vrsti klasifikatorjev: Class-0, pri kateri razredne oznake neposredno služijo kot izhod, in Class-1, ki uporablja regresijo za določitev lokalnega linearnega modela v posledičnem delu. Primer slednjega je model pClass~\cite{Pratama_Anavatti_Joo_Lughofer_2015}, ki uporablja Gaussove elipsoide za definicijo rojev. Podobno kot naš pristop uporablja koncept volumna rojev, vendar pri izračunu pripadnosti vzorca in mehanizmu odstranjevanja pravil. Podobno je bil v~\cite{Lughofer_2022} predstavljen EFC, ki uporablja nenadzorovano rojenje za združevanje podatkov in sprotno aktivno učenje za izbiro najbolj informativnih vzorcev za označevanje s strani strokovnjakov. Ta metoda omogoča več uporabnikom, da označijo svoje podatke in ustvarijo klasifikatorje, ki so nato združeni v ansambel na podlagi konsenza.
\par Mehki modeli so bili že uporabljeni v federativnem učenju za zelo različne namene. Na primer, za klasifikacijo so v~\cite{Poap_2021} bili usposobljeni različni modeli federativnega učenja, pri čemer je bil uporabljen mehki model za določitev izbire modela. V~\cite{Zhang_Shi_Chang_Lin_2023} je bila uvedena porazdeljena mehka nevronska mreža, ki se spopada z nehomogenimi in negotovimi podatki z uporabo samorazvijajočega se mehanizma za dodajanje novih pravil in deaktivacijo obstoječih. Prav tako zagotavlja izbiro podnabora pravil za vsakega lastnika, kar je zelo pomembno v federativnem učenju. Poleg tega so bile mehke nevronske mreže nedavno uporabljene v porazdeljenem nenadzorovanem učenju v~\cite{Shi_Lin_Chang_Ding_Shi_Yao_2021}, ki uporablja rojenje s K-središči (\textit{K-means}), in v porazdeljenem pol-nadzorovanem učenju. Podobno je bila v~\cite{Stallmann_Wilbik_2022} predlagana federativna metoda mehkega rojenja s $c$-središči (\textit{fuzzy $c$-means} FCM) za nenadzorovano rojenje, da bi se spopadli z nehomogenimi podatki. Zanimivo je, da avtorji slednje študije poudarjajo potrebo po izbiri števila združkov kot nerešen problem. Slabost teh metod je uporaba osno vzporednih Gaussovih rojev za definiranje njihovih antecedentov, ki so manj informativni kot elipsoidna predstavitev~\cite{Pratama_Anavatti_Joo_Lughofer_2015}. Še bolj ključna težava pa je predpostavka o številu rojev, ki ne omogoča dodajanja novih rojev, če se ti pojavijo po fazi učenja.
\section{Metodologija}
\par Osnovni gradniki samorazvijajočega se mehkega modela so Gaussovi roji, definirani s središčem $\tensor{\mu}_i \,{\in}\, \R^\D$, kovariančno matriko $\tensor{\Sigma}_i {\in} \R^{\D \times \D}$ in številom vzorcev $n_i \,{\in}\, \N$, kjer je $i \,{\in}\, \mathcal{I}, \mathcal{I} \,{=}\, {1,...,c}$ indeks roja. Za izračun razdalje med vzorcem in vsakim rojem uporabljamo Mahalanobisovo razdaljo $d_i^2 \,{\in}\, \R^+$, s katero nato izračunamo pripadnostne funkcije $\gamma_i \,{\in}\, [0,1]$ vzorca vsakemu pravilu:
\begin{equation}
\gamma_i = \mathrm{exp}\Big({-}\frac{1}{\D}(\tensor{x}{-}\tensor{\mu}_i)^{\top}\tensor{\Sigma}_i^{-1}(\tensor{x}{-}\tensor{\mu}_i)\Big).
\end{equation}
Pri klasifikaciji imamo podatkovno zbirko z $M$ razredi, kjer za vsak razred naučimo svoj klasifikator (One-Vs-All). Vsako mehko pravilo je sestavljeno iz roja in posledičnega dela, ki vsebuje kodiranje razreda \( \tensor{\theta}_i  \,{\in}\, \{0,1\}^M \). To kodiranje je $M$-dimenzionalni binarni vektor, kjer vsaka dimenzija edinstveno predstavlja razred. V sprotnem načinu delovnaja sistem ocenjuje varianco podatkov za vsak razred in izbere najbolj informativne značilke s pragom $\kappa_F$ Fisherjeve ocene (\textit{Fisher's score}), ki je mera prekrivanja distribuciji. Izhod klasifikatorja ustreza roju z najvišjo pripadnostjo za vzorec~\cite{Angelov_Xiaowei_Zhou_2008, Pratama_Anavatti_Joo_Lughofer_2015}:
\begin{equation} \label{eq:hard_output}
\tensor{\hat{y}}(\tensor{x}) = \tensor{\theta}_{i^*}, \quad \text{kjer je} \quad i^* = \underset{i  \,{\in}\, \I}{\mathrm{argmax}}\ \gamma_i(\tensor{x}).
\end{equation}
Vzorec je mogoče opisati z obstoječimi mehkimi pravili, če je pogoj $ \gamma_{i^*}\,{>}\, \exp({-\mathrm{N}_\sigma^2}/{\D})$ izpolnjen. Parametri roja ${i^*}$ se inkrementalno naučijo z enačbami~\cite{Skrjanc_2020}:
\begin{align}
\tensor{e}_{i^*}&= \tensor{x} - \tensor{\mu}_{i^*}, \label{eq:incremental_clustering_1} \\
\tensor{\mu}_{i^*}&= \tensor{\mu}_{i^*}+ {\tensor{e}_{i^*}}/{(n_{i^*}{+}1)}, \label{eq:incremental_clustering_2} \\
\tensor{S}_{i^*}&= \tensor{S}_{i^*}+ \tensor{e}_{i^*}\big(\tensor{x} {-} \tensor{\mu}_{i^*}\big)^\top, \label{eq:incremental_clustering_3} \\
n_{i^*}&= n_{i^*}+1, \label{eq:incremental_clustering_4}
\end{align}
Če pa vzorca ni mogoče opisati z obstoječimi pravili, se doda novo pravilo ${i^*}\,{=}\,c\,{+}\,1$ s središčem v $\tensor{\mu}_{i^*} \,{=} \,\tensor{x}$, $n_{i^*}\,{=}\,1$ in $
\tensor{\Sigma}_{i^*} \,{=}\, \mathrm{diag}\left({\tensor{\sigma}^2}/{\mathrm{N}_r}\right)$, kjer je $\tensor{\sigma}^2  \,{\in}\, R^\D$ ocena variance vseh podatkov in $\mathrm{N}_r \,{\in}\, \mathbb{R}^+$ kvantizacijsko število, ki vpliva na prevzeto velikost rojev. Vrednost aktivacije smo nastavili na $\mathrm{N}_\sigma {=} \sqrt{\D}$. 
\par Med sprotnim učenjem lahko nastane večje število prekrivajočih se rojev, zato samorazvijajoči se mehki sistemi uporabljajo mehanizem združevanja rojev. Mehanizem združi pare rojev $p, q  \,{\in}\, \left\{i \mid \gamma_i \,{>}\, \exp(-\mathrm{N}_\sigma^2/\D)\right\}$, ki izpolnjujejo pogoj prekrivanja~\cite{Skrjanc_2020}: 
\begin{math} \label{eq:volume_condition}
    {V_{pq}}/{(V_p {+} V_q)}  < \kappa_m^\D\end{math}, kjer $pq$ označuje združen roj in $V_i {=}\frac{2 \pi^{\D / 2}}{d \Gamma(\D / 2)} {|\tensor{\Sigma}_i|}$ je prostornina hiperelipse, ki predstavlja roj. Ta pogoj izračunamo za vse kombinacije rojev paralelno. Središče in število vzorcev združenega roja izračunamo kot~\cite{Skrjanc_2020}:
\begin{align}
    \tensor{\mu}_{pq} = \frac{n_p\tensor{\mu}_{p} {+} n_q\tensor{\mu}_{q}}{n_{pq}}, \, \, \text{kjer je}  \,\,  n_{pq} = n_p {+} n_q, \label{eq:merge_samples} \quad
\end{align}
in novo kovariančno matriko $\Sigma_{pq}$ pa kot:
\begin{equation}\label{eq:merging_equation}
\begin{split}
{(n_{pq}{-}1)}\tensor{\Sigma}_{pq}  = \, & (n_{p} {-} 1)\tensor{\Sigma}_p +{(n_{q} {-} 1)}\tensor{\Sigma}_q +\\ & + \frac{n_{p}n_{q}}{n_{pq}} (\tensor{\mu}_{p} {-} \tensor{\mu}_{q})(\tensor{\mu}_{p} {-} \tensor{\mu}_{q})^\top.
 \end{split}
\end{equation}
Ta zapis omogoča združevanje rojev brez shranjevanja vzorcev posameznega roja, kar omogoča združevanje rojev na strežniku, kjer vzorci niso na voljo. Ta izračun se izvaja za vse kombinacije kandidatnih rojev hkrati. Ko so kovariančne matrike določene, se izračuna volumenski pogoj, in združijo se najbolj primerni kandidati. Eden od problemov, ki se pojavi pri združevanju rojev, je, da lahko en roj postane prevelik in pogoltne vse druge~\cite{Pratama_Anavatti_Joo_Lughofer_2015}. Da bi se temu izognili, smo pri združevanju omejili velikost rojev z večkratnikom volumna prototipnega roja.
\par Po učenju s privatnimi podatki, lastniki podatkov posredujejo parametre svojih modelov na strežnik, kjer se vsi lokalni modeli dodajo h globalnemu modelu in izvede mehanizem združevanja. Sistem izbere kandidate za združevanje tako, da za vsa središča rojev izračuna razdalje do ostalih rojev. Poleg mehanizma združevanja pravil uporabljamo tudi metodo odstranjevanja na podlagi starosti rojev, ki je definirana kot čas od zadnje aktivacije pravila med učenjem. Po združevanju na strežniku se odstranijo najstarejša pravila, kar omogoča, da sistem odstrani osamelce in ohrani najbolj relevantna pravila. Globalni model se nato prenese nazaj k lokalnim vozliščem.
%

%V 
\begin{figure*}[!ht]
    \centering
    \includegraphics[width=0.9\linewidth]{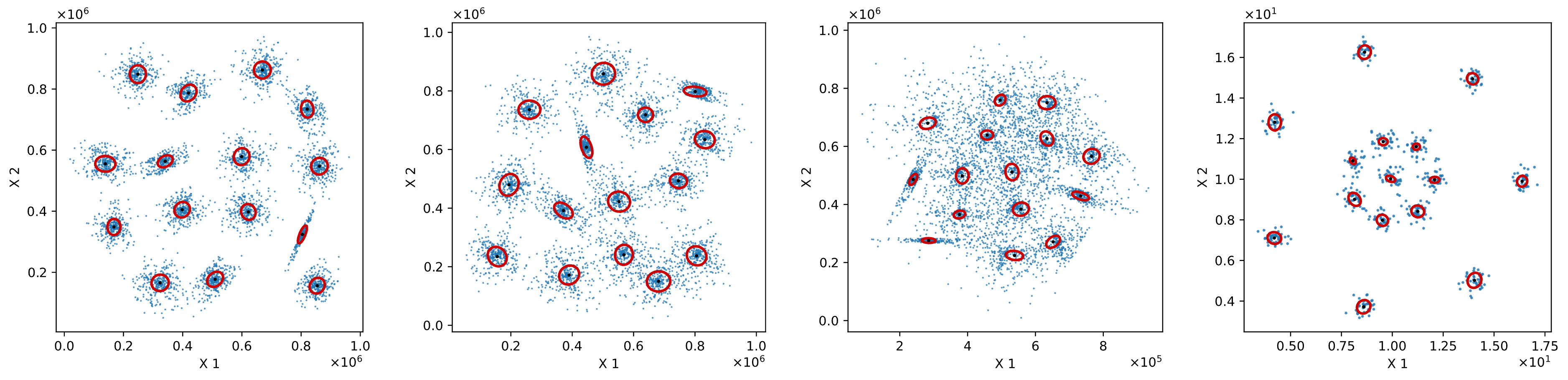}
    \includegraphics[width=0.9\linewidth]{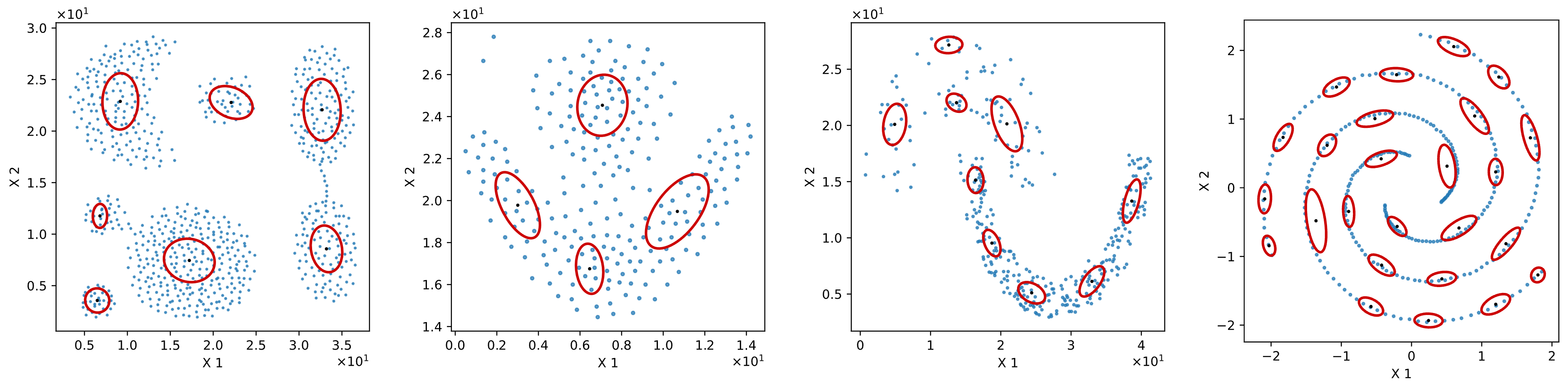}
    \caption{Federativno rojenje naborov sintetičnih podatkov z Gaussovo distribucijo (zgoraj od leve proti desni): S1, S2, S4, R15; in naborov podatkov z ne-Gaussovo distribucijo (spodaj od leve proti desni): Aggregation, Flame, Jain, Spiral. Prikazani so podatki (modro) in 2$\sigma$ elipse, ki predstavljajo roje (rdeče).}
    \label{fig:clustering}
\end{figure*}

\begin{table*}[!ht]
\centering
\setlength{\tabcolsep}{2pt}
\scriptsize
\begin{tabular}{@{}ll|ccccccc|c@{}}
\toprule
\multirow{2}{*}{Podatkovna zbirka} & \multirow{2}{*}{Mere} & \multicolumn{7}{c}{Metode} \\
\cmidrule{3-10}
& & XGBoost & Logistic & Naive & KNN & SVM(RBF) & Decision & ALMMo-1 & Naša\\
& & & Regression & Bayes & & & Tree & \\
\midrule %Iris
\multirow{4}{*}{\shortstack[l]{Perunika \\ (Iris)}} & Natančnost [\%] & 94.1${\pm}$2.5& 95.1${\pm}$2.2& 95.2${\pm}$2.0& 94.9${\pm}$2.0& 95.7${\pm}$2.2& 93.9${\pm}$3.0& 66.7${\pm}$6.3& \textbf{96.5${\pm}$1.8}\\
                           & F1 score [\%] & 94.0${\pm}$2.5& 95.1${\pm}$2.2& 95.2${\pm}$2.0& 94.9${\pm}$2.0& 95.7${\pm}$2.2& 93.9${\pm}$3.0& 55.8${\pm}$7.5& \textbf{96.5${\pm}$1.8}\\
                           & ROC AUC [\%] & 98.2${\pm}$1.1& 99.7${\pm}$0.3& 99.5${\pm}$0.4& 99.0${\pm}$1.0& 99.8${\pm}$0.2& 95.5${\pm}$2.3& 50.0${\pm}$0.0& \textbf{99.9${\pm}$0.2}\\                           & Čas / vzorec [ms] & 0.24${\pm}$0.05& 0.01${\pm}$0.02& \textbf{0.00${\pm}$0.00}& 0.01${\pm}$0.02& 0.01${\pm}$0.00& 0.00${\pm}$0.00& 0.76${\pm}$0.11& 1.44${\pm}$0.08\\
\midrule %Wine
\multirow{4}{*}{\shortstack[l]{Vino \\ (Wine) }} & Natančnost [\%] & 96.7${\pm}$1.8& 98.0${\pm}$1.6& 97.5${\pm}$1.5& 95.6${\pm}$2.2& 98.2${\pm}$1.4& 90.3${\pm}$3.9& 68.5${\pm}$3.7& \textbf{98.3${\pm}$1.6}\\
                           & F1 score [\%] & 96.7${\pm}$1.8& 98.0${\pm}$1.7& 97.5${\pm}$1.5& 95.6${\pm}$2.3& 98.2${\pm}$1.4& 90.3${\pm}$3.9& 59.1${\pm}$4.2& \textbf{98.3${\pm}$1.6}\\
                           & ROC AUC [\%] & 99.7${\pm}$0.3& \textbf{100.0${\pm}$0.1}& 99.9${\pm}$0.2& 99.4${\pm}$0.8& 100.0${\pm}$0.1& 92.6${\pm}$3.0& 50.0${\pm}$0.0& 99.9${\pm}$0.1\\
                           & Čas / vzorec [ms] & 0.18${\pm}$0.03& 0.01${\pm}$0.01& \textbf{0.00${\pm}$0.00}& 0.01${\pm}$0.01& 0.01${\pm}$0.00& 0.00${\pm}$0.00& 0.72${\pm}$0.10& 1.28${\pm}$0.05\\
                           
\midrule %Heart Disease
\multirow{4}{*}{\shortstack[l]{Bolezni srca \\ (Heart disease) }} & Natančnost [\%] & 79.0${\pm}$3.7& 83.2${\pm}$4.2& \textbf{83.5${\pm}$3.7}& 82.0${\pm}$3.3& 82.3${\pm}$3.7& 72.4${\pm}$4.4& \textbf{83.5${\pm}$3.8}& 83.1${\pm}$3.8\\
                           & F1 score [\%] & 79.0${\pm}$3.7& 83.1${\pm}$4.3& \textbf{83.4${\pm}$3.7}& 82.0${\pm}$3.3& 82.2${\pm}$3.7& 72.4${\pm}$4.3& 83.4${\pm}$3.9& 81.6${\pm}$3.7\\
                           & ROC AUC [\%] & 87.7${\pm}$3.0& \textbf{90.2${\pm}$2.7}& 89.6${\pm}$2.6& 88.5${\pm}$2.9& 89.6${\pm}$2.7& 72.5${\pm}$4.3& 50.0${\pm}$0.0& 89.4${\pm}$2.5\\
                           & Čas / vzorec [ms] & 0.09${\pm}$0.03& 0.00${\pm}$0.01& \textbf{0.00${\pm}$0.00}& 0.01${\pm}$0.02& 0.01${\pm}$0.01& 0.00${\pm}$0.01& 0.69${\pm}$0.14& 1.13${\pm}$0.08\\
                                                     
\midrule %Brest cancer
\multirow{4}{*}{\shortstack[l]{Rak dojke \\ (Brest cancer)}} & Natančnost [\%] & 96.2${\pm}$1.2& \textbf{97.7${\pm}$0.8}& 93.3${\pm}$1.4& 96.4${\pm}$1.0& 97.4${\pm}$0.8& 92.6${\pm}$2.1& 95.6${\pm}$1.2& 96.4${\pm}$1.5\\
                           & F1 score [\%] & 96.2${\pm}$1.1& \textbf{97.7${\pm}$0.8}& 93.3${\pm}$1.4& 96.3${\pm}$1.0& 97.4${\pm}$0.8& 92.6${\pm}$2.1& 95.5${\pm}$1.2& 95.2${\pm}$1.9\\
                           & ROC AUC [\%] & 99.1${\pm}$0.5& 99.5${\pm}$0.3& 98.5${\pm}$0.6& 98.4${\pm}$0.6& \textbf{99.5${\pm}$0.3}& 92.2${\pm}$2.2& 50.0${\pm}$0.0& 99.4${\pm}$0.5\\
                           & Čas / vzorec [ms] & 0.06${\pm}$0.02& 0.00${\pm}$0.00& \textbf{0.00${\pm}$0.00}& 0.02${\pm}$0.01& 0.01${\pm}$0.01& 0.01${\pm}$0.01& 0.75${\pm}$0.13& 1.02${\pm}$0.06\\
\midrule %MNIST $8\times8$
\multirow{4}{*}{\shortstack[l]{Številke \\ (Digits) }} & Natančnost [\%] & 96.1${\pm}$1.0& 96.7${\pm}$0.7& 78.5${\pm}$3.6& 97.1${\pm}$0.7& 98.0${\pm}$0.5& 84.5${\pm}$1.5& 11.5${\pm}$1.2& \textbf{98.1${\pm}$0.4}\\
                           & F1 score [\%] & 96.1${\pm}$0.9& 96.7${\pm}$0.7& 78.4${\pm}$3.7& 97.1${\pm}$0.7& 98.0${\pm}$0.5& 84.5${\pm}$1.5& 4.3${\pm}$0.7& \textbf{98.1${\pm}$0.4}\\
                           & ROC AUC [\%] & 99.9${\pm}$0.0& 99.9${\pm}$0.0& 97.2${\pm}$0.4& 99.6${\pm}$0.1& \textbf{99.9${\pm}$0.0}& 91.4${\pm}$0.8& 50.0${\pm}$0.0& 99.8${\pm}$0.1\\
                           & Čas / vzorec [ms] & 0.13${\pm}$0.02& 0.01${\pm}$0.00& \textbf{0.00${\pm}$0.00}& 0.02${\pm}$0.00& 0.08${\pm}$0.00& 0.00${\pm}$0.00& 1.25${\pm}$0.15& 1.45${\pm}$0.10\\
%\midrule 
% \multirow{4}{*}{\shortstack[l]{Srčno popuščanje \\ (Heart failure)}} & Natančnost [\%] & \textbf{82.8${\pm}$3.0}& 81.8${\pm}$3.6& 76.7${\pm}$4.2& 74.1${\pm}$3.9& 80.0${\pm}$3.6& 77.6${\pm}$3.6& 81.4${\pm}$3.1& 80.7${\pm}$4.2\\
%                            & F1 score [\%] & \textbf{82.6${\pm}$3.1}& 81.4${\pm}$3.7& 75.1${\pm}$5.1& 71.3${\pm}$4.9& 79.3${\pm}$3.9& 77.5${\pm}$3.8& 80.7${\pm}$3.4& 67.8${\pm}$6.2\\
%                            & ROC AUC [\%] & \textbf{89.2${\pm}$2.7}& 86.4${\pm}$3.0& 83.9${\pm}$3.5& 74.5${\pm}$4.8& 85.4${\pm}$2.8& 74.4${\pm}$4.9& 50.0${\pm}$0.0& 82.2${\pm}$5.3\\
%                            & Čas / vzorec [ms] & 0.10${\pm}$0.04& 0.00${\pm}$0.01& 0.00${\pm}$0.00& 0.01${\pm}$0.01& 0.01${\pm}$0.02& \textbf{0.00${\pm}$0.00}& 0.48${\pm}$0.12& 1.01${\pm}$0.09\\
                            %  Autism
                           \midrule
\multirow{4}{*}{\shortstack[l]{Avtizem \\ (Autism)}} & Natančnost [\%] & \textbf{100.0${\pm}$0.0}& \textbf{100.0${\pm}$0.0}& 96.6${\pm}$1.1& 96.1${\pm}$1.2& 99.2${\pm}$1.1& \textbf{100.0${\pm}$0.0}& 95.5${\pm}$1.4& \textbf{100.0${\pm}$0.0}\\
                           & F1 score [\%] & \textbf{100.0${\pm}$0.0}& \textbf{100.0${\pm}$0.0}& 96.6${\pm}$1.1& 96.1${\pm}$1.2& 99.2${\pm}$1.1& \textbf{100.0${\pm}$0.0}& 95.5${\pm}$1.4& \textbf{100.0${\pm}$0.0}\\
                           & ROC AUC [\%] & \textbf{100.0${\pm}$0.0}& \textbf{100.0${\pm}$0.0}& 99.1${\pm}$0.9& 99.3${\pm}$0.4& 100.0${\pm}$0.1& \textbf{100.0${\pm}$0.0}& 50.0${\pm}$0.0& \textbf{100.0${\pm}$0.0}\\
                           & Čas / vzorec [ms] & 0.03${\pm}$0.01& 0.01${\pm}$0.01& 0.00${\pm}$0.00& 0.02${\pm}$0.01& 0.01${\pm}$0.01& \textbf{0.00${\pm}$0.00}& 0.97${\pm}$0.18& 0.56${\pm}$0.10\\
 
% \midrule %NSL-KDD
% \multirow{4}{*}{\shortstack[l]{Kibernetska varnost \\ (NSL-KDD)}} & Natančnost [\%] & \textbf{99.6${\pm}$0.1}& 98.5${\pm}$0.2& 90.2${\pm}$1.3& 99.5${\pm}$0.0& 99.4${\pm}$0.1& 99.4${\pm}$0.1& 97.2${\pm}$0.2& 98.7${\pm}$0.2\\
%                            & F1 score [\%] & \textbf{99.6${\pm}$0.1}& 98.5${\pm}$0.2& 90.2${\pm}$1.3& 99.5${\pm}$0.0& 99.4${\pm}$0.1& 99.4${\pm}$0.1& 97.2${\pm}$0.2& 98.8${\pm}$0.2\\
%                            & ROC AUC [\%] & \textbf{100.0${\pm}$0.0}& 99.7${\pm}$0.0& 95.9${\pm}$0.7& 99.9${\pm}$0.0& 100.0${\pm}$0.0& 99.4${\pm}$0.1& 50.0${\pm}$0.0& 99.5${\pm}$0.1\\
%                            & Čas / vzorec [ms] & 0.01${\pm}$0.00& 0.00${\pm}$0.00& \textbf{0.00${\pm}$0.00}& 0.01${\pm}$0.00& 0.05${\pm}$0.00& 0.01${\pm}$0.00& 0.91${\pm}$0.07& 1.30${\pm}$0.09\\
\bottomrule
\end{tabular}
\caption{Primerjava predlagane metode za problem klasifikacije z različnimi nabori podatkov. Predstavljena je povprečna vrednost $\pm$ standardna deviacija natančnosti (\textit{accuracy}), F1 ocena (\textit{F1 score}), površina pod krivuljo operacijskih karakteristik sprejemnika (\textit{Receiver Operating Characteristic Area Under the Curve} -- {ROC AUC}) in čas učenja za posamezen vzorec v milisekundah.}
\label{tab:klasifikaciija}
\end{table*}
\section{Eksperimentiranje} 
\par Izvedli smo eksperimente federativnega rojenja in klasifikacije s predlagano metodo. Metoda je bila implementirana s knjižnico PyTorch, ki omogoča delovanje na procesorju ali grafični kartici. Rojenje je ključen del predstavljene metode, ne glede na to, ali je naš cilj rojenje ali klasifikacija. Rojenje smo izvedli na 2D podatkovnih zbirkah\footnote{https://cs.uef.fi/sipu/datasets} z Gaussovo porazdelitvijo procesa in podatkovnih zbirkah arbitrarne porazdelitve. Podatki so bili naključno porazdeljeni med 3 lastnike. Vsak udeleženec zgradi svoj lokalni model. Modeli se nato agregirajo na strežniku v eni rundi komunikacije. Načrtovalske parametre moramo nekoliko prilagoditi vsaki nalogi in podatkovni zbirki. Pri tem smo nastavili parametre za vse zbirke podatkov na $\kappa_n {=} \mathrm{N}_r$ in $\kappa_m {=} 1.5$. Parameter $N_r$ (kvantizacijo prostora) smo določili empirično za vsako zbirko podatkov.
\par Drugi del raziskave se osredotoči na razširitev rojenja v uporabo metode rojenja za klasifikacijo. Predlagano metodo smo primerjali z uveljavljenimi metodami klasifikacije: XGBoost, logistična regresija, naivni Bayes, k-najbližjih sosedov (\textit{k-Nearest Neighbors} -- KNN), podporno vektorski stroj (\textit{Support Vector Machine} -- SVM), odločitveno drevo (Decision tree) ter samorazvijajoč se mehki klasifikator ALMMo-1~\cite{Angelov_Gu_Principe_2018}. Uporabljene podatkovne zbirke so na voljo v repozitoriju UCI\footnote{https://archive.ics.uci.edu/datasets}. Najmanjše število vzorcev smo nastavili na $\kappa_n {=} 1$ in druge parametre metode smo empirično prilagodili vsaki zbirki podatkov. Za našo metodo smo učne podatke razdelili med 3 lastnike, medtem ko ostale metode niso federativne. Eksperiment smo izvedli s 3-kratno navzkrižno validacijo (\textit{K-fold}) in eksperimente ponovili 10-krat.

\section{Rezultati in diskusija}
\par Rezultati rojenja so grafično prikazani na sliki \ref{fig:clustering}. Predlagana metoda zelo dobro opiše podatke, ki temeljijo na Gaussovih procesih. Je pa nekoliko manj robustna na prekrivajoče se podatke in zaradi metode združevanja rojev ne omogoča koncentričnih rojev s podobnimi lastnimi vektorji. V primeru arbitrarne porazdelitve podatkov posamezni Gaussovi roji ne morejo v celoti opisati dejanskih rojev. Poglavitna prednost te metode je, da ni občutljiva na število vzorcev, glej S4. Primerjava prostornin dveh rojev postane nezanesljiva v višjih dimenzijah, ker je prostornina kovariančne matrike določena s produktom lastnih vrednosti. Stalno večje ali manjše vrednosti lastnih vrednosti vodijo do velike razlike v prostorninah, tudi če se zdi, da imajo roji približno enako obliko v vseh dimenzijah. Pri tem eksperimentu smo spreminjali le kvantizacijo prostora, zato se pojavi vprašanje, ali je mogoče ta parameter izbrati avtomatsko med delovanjem glede na velikost najmanjšega roja.

\par Rezultati klasifikacije so predstavljeni v tabeli \ref{tab:klasifikaciija}. Vidimo, da je predlagana metoda dosegla najboljšo natančnost na večini izbranih podatkovnih zbirk ali pa je bila primerljiva z ostalimi metodami. Metoda ALMMo sicer omogoča klasifikacijo več razredov, vendar najbolje deluje pri binarni klasifikaciji. Naša metoda izkazuje boljšo natančnost pri numeričnih značilkah kot pri kategorialnih značilkah, kljub uporabi kodiranja. Žal pa je naša metoda precej počasnejša od primerjanih metod, ki so implementirane v dobro optimiziranih knjižnicah. Večina računskega časa naše metode je posledica izračuna pogoja prekrivanja v zanki, ki vključuje izračun več determinant. Ta izračun je potrebno izvesti za več kombinacij rojev, da najdemo najprimernejšega za združevanje. Mehanizem združevanja se sicer izračuna za vse pare rojev hkrati, vendar metoda temelji na sprotnem učenju, ki obravnava vsak vzorec zaporedno. Grafične kartice delujejo hitro, če lahko izvedemo veliko število operacij vzporedno in brez shranjevanja vmesnih vrednosti, kar je v nasprotju s sprotnim učenjem. Metoda je hitrejša na procesorju za manjše zbirke podatkov, kot so Iris in Wine, vendar hitrejša na grafični kartici za ostale večje zbirke podatkov. 
\section{Zaključek}
\par V raziskavi smo prilagodili metodologijo, razvito za identifikacijo sistemov iz razvijajočih se podatkovnih tokov, in jo uporabili za federativno učenje. Ključna prednost tega pristopa je, da se ne zanaša na vnaprejšnje znanje o številu rojev in da so lokalni modeli neposredno vključeni v globalni model. Poleg tega se sistem lahko spreminja skozi čas in v primeru, da se število pričakovanih rojev spremeni. Glavna omejitev metode je njena omejena zmožnost paralelnega računanja; čeprav je bila večina delov metode vektorizirana, inkrementalno rojenje ni bilo. Združevanje rojev je časovno najbolj zahteven del metode. Združevanje rojev je preprosto, ampak mera prekrivanja vključuje izračun determinante vseh kombinacij rojev, ki jih želimo združiti. Pokazali smo, da predlagana metoda doseže primerljive rezultate z uveljavljenimi klasifikatorji na standardnih naborih podatkov. Nadaljnje delo vključuje primerjavo metode s federativnimi metodami rojenja in klasifikacije. Velik potencial ima delno nadzorovano učenje, ki vključuje tako označene kot neoznačene podatke.
\small

% \bibitem{ERK} ERK, http://www.ieee.si/erk/index.html 
% \bibitem{Zbornik} B. Zajc, A. Trost: Zbornik triindvajsete mednarodne Elektrotehniške in računalniške konference ERK 2014, 22. - 24. September 2014, Portorož, Slovenija

% \end{thebibliography}

\bibliographystyle{IEEEtran}
%\bibliography{./ref.bib}
\input{erk.bbl}

\end{document}

%% file: erk.bbl

%% file: erk.bbl
\begin{thebibliography}{10}
\providecommand{\url}[1]{#1}
\csname url@samestyle\endcsname
\providecommand{\newblock}{\relax}
\providecommand{\bibinfo}[2]{#2}
\providecommand{\BIBentrySTDinterwordspacing}{\spaceskip=0pt\relax}
\providecommand{\BIBentryALTinterwordstretchfactor}{4}
\providecommand{\BIBentryALTinterwordspacing}{\spaceskip=\fontdimen2\font plus
\BIBentryALTinterwordstretchfactor\fontdimen3\font minus \fontdimen4\font\relax}
\providecommand{\BIBforeignlanguage}[2]{{%
\expandafter\ifx\csname l@#1\endcsname\relax
\typeout{** WARNING: IEEEtran.bst: No hyphenation pattern has been}%
\typeout{** loaded for the language `#1'. Using the pattern for}%
\typeout{** the default language instead.}%
\else
\language=\csname l@#1\endcsname
\fi
#2}}
\providecommand{\BIBdecl}{\relax}
\BIBdecl

\bibitem{McMahan_Moore_Ramage_Hampson_Arcas_2023}
H.~B. McMahan, E.~Moore, D.~Ramage, S.~Hampson, and B.~A.~y. Arcas, ``{Communication-Efficient Learning of Deep Networks from Decentralized Data},'' no. arXiv:1602.05629, Jan. 2023.

\bibitem{Konecnx_McMahan_2017}
J.~Konečný, H.~B. McMahan, F.~X. Yu, P.~Richtárik, A.~T. Suresh, and D.~Bacon, ``{Federated Learning: Strategies for Improving Communication Efficiency},'' no. arXiv:1610.05492, Oct. 2017.

\bibitem{Angelov_Xiaowei_Zhou_2008}
P.~Angelov and X.~Zhou, ``\BIBforeignlanguage{en}{{Evolving Fuzzy-Rule-Based Classifiers From Data Streams}},'' \emph{\BIBforeignlanguage{en}{IEEE Transactions on Fuzzy Systems}}, vol.~16, no.~6, p. 1462–1475, Dec. 2008.

\bibitem{Pratama_Anavatti_Joo_Lughofer_2015}
M.~Pratama, S.~G. Anavatti, M.~Joo, and E.~D. Lughofer, ``\BIBforeignlanguage{en}{{pClass: An Effective Classifier for Streaming Examples}},'' \emph{\BIBforeignlanguage{en}{IEEE Transactions on Fuzzy Systems}}, vol.~23, no.~2, p. 369–386, Apr. 2015.

\bibitem{Skrjanc_2020}
I.~Škrjanc, ``{Cluster-Volume-Based Merging Approach for Incrementally Evolving Fuzzy Gaussian Clustering-eGAUSS+},'' \emph{IEEE Transactions on Fuzzy Systems}, vol.~28, no.~9, p. 2222–2231, Sep. 2020.

\bibitem{Lughofer_2022}
E.~Lughofer, ``{Evolving multi-user fuzzy classifier systems integrating human uncertainty and expert knowledge},'' \emph{Information Sciences}, vol. 596, p. 30–52, Jun. 2022.

\bibitem{Stallmann_Wilbik_2022}
M.~Stallmann and A.~Wilbik, ``\BIBforeignlanguage{en}{{On a Framework for Federated Cluster Analysis}},'' \emph{\BIBforeignlanguage{en}{Applied Sciences}}, vol.~12, no. 2020, p. 10455, Jan. 2022.

\bibitem{Li_Huang_Yang_Wang_Zhang_2020}
X.~Li, K.~Huang, W.~Yang, S.~Wang, and Z.~Zhang, ``{On the Convergence of FedAvg on Non-IID Data},'' no. arXiv:1907.02189, Jun. 2020.

\bibitem{Poap_2021}
D.~Poap, ``\BIBforeignlanguage{en}{{Fuzzy Consensus With Federated Learning Method in Medical Systems}},'' \emph{\BIBforeignlanguage{en}{IEEE Access}}, vol.~9, p. 150383–150392, 2021.

\bibitem{Zhang_Shi_Chang_Lin_2023}
L.~Zhang, Y.~Shi, Y.-C. Chang, and C.-T. Lin, ``{Federated Fuzzy Neural Network with Evolutionary Rule Learning},'' \emph{IEEE Transactions on Fuzzy Systems}, vol.~31, no.~5, p. 1653–1664, May 2023.

\bibitem{Shi_Lin_Chang_Ding_Shi_Yao_2021}
Y.~Shi, C.-T. Lin, Y.-C. Chang, W.~Ding, Y.~Shi, and X.~Yao, ``{Consensus Learning for Distributed Fuzzy Neural Network in Big Data Environment},'' \emph{IEEE Transactions on Emerging Topics in Computational Intelligence}, vol.~5, no.~1, p. 29–41, Feb. 2021.

\bibitem{Angelov_Gu_Principe_2018}
P.~P. Angelov, X.~Gu, and J.~C. Principe, ``Autonomous learning multimodel systems from data streams,'' \emph{IEEE Transactions on Fuzzy Systems}, vol.~26, no.~4, p. 2213–2224, Aug. 2018.

\end{thebibliography}
